\documentclass[preprint,12pt]{elsarticle}

\usepackage{amssymb}
\usepackage{amsmath}
\usepackage{algorithm}
\usepackage{algorithmic}
\usepackage{multirow}
\usepackage{graphicx}
\usepackage{subfig}

\usepackage{verbatim}
\usepackage{array}
\usepackage{textcomp}
\usepackage{stfloats}
\usepackage{float}

\journal{Eng. Applications of Artificial Intelligence}

\begin{document}

\begin{frontmatter}



\title{A generalised novel loss function for computational fluid dynamics}


\author{Zachary Cooper-Baldock\corref{cor1}\fnref{label1}}
\author[label1]{Paulo E. Santos} 
\author[label1]{Russell S.A. Brinkworth} 
\author[label1]{Karl Sammut} 

\affiliation[label1]{organization={Flinders University, Centre for Defence Engineering, Research and Training},
            city={Adelaide},
            postcode={5042}, 
            state={South Australia},
            country={Australia}}
\cortext[cor1]{Corresponding author: zachary.cooperbaldock@flinders.edu.au}

\small
\begin{abstract} Computational fluid dynamics (CFD) simulations are crucial in automotive, aerospace, maritime and medical applications, but are limited by the complexity, cost and computational requirements of directly calculating the flow, often taking days of compute time. Machine-learning architectures, such as controlled generative adversarial networks (cGANs) hold significant potential in enhancing or replacing CFD investigations, due to cGANs ability to approximate the underlying data distribution of a dataset. Unlike traditional cGAN applications, where the entire image carries information, CFD data contains small regions of highly variant data, immersed in a large context of low variance that is of minimal importance. This renders most existing deep learning techniques that give equal importance to every portion of the data during training, inefficient. To mitigate this, a novel loss function is proposed called Gradient Mean Squared Error (GMSE) which automatically and dynamically identifies the regions of importance on a field-by-field basis, assigning appropriate weights according to the local variance. To assess the effectiveness of the proposed solution, three identical networks were trained; optimised with Mean Squared Error (MSE) loss, proposed GMSE loss and a dynamic variant of GMSE (DGMSE). The novel loss function resulted in faster loss convergence, correlating to reduced training time, whilst also displaying an 83.6\% reduction in structural similarity error between the generated field and ground truth simulations, a 76.6\% higher maximum rate of loss and an increased ability to fool a discriminator network. It is hoped that this loss function will enable accelerated machine learning within computational fluid dynamics.
\end{abstract}



\end{frontmatter}

\section{Introduction}
Computational fluid dynamics (CFD) data developed from automotive, aerospace, maritime and medical applications diverges markedly from traditional images. This is due to the fact that CFD studies the movements of fluids around or through objects of interest. CFD techniques then result in the generation of gradient containing fields that offer insights into the physical changes to the velocity, density and pressure of the fluid. These produced fields, both in 2D and 3D forms, do not conform to traditional visual patterns seen in images \cite{MNIST_Data}, such as faces \cite{GANforFace}, text and numbers \cite{GANforNumbers} or more general images \cite{GANforGeneral}, upon which the focus of machine learning often sits. The gradients in CFD flow fields are frequently irregular and influenced by various fluid specific effects \cite{CompFluids1}. This results in CFD data containing critical components that manifest at a small scale within a much larger flow field, where the majority of the field carries minimal information. Effectively adapting and translating generative machine learning architectures for CFD requires a more nuanced approach \cite{CompFluids2} to isolate and evaluate predictions in these intricate regions. Historically, this has seen the development of highly specific, tailored loss functions \cite{CompFluids2, CFD_loss_1, CFD_loss_2, CFD_loss_3}. Where, in machine learning, a loss function serves to map a change in performance to a numerical value, often representing a performance metric. It is employed to assess the learned behaviour of a network, architecture, algorithm, or system of algorithms. The resulting value from the loss function is then used for optimization, where the objective is to decrease or increase the value of the function, based on the specific task. 
\\ \\
There are two common approaches to loss function development for a deep learning architecture; generalised or specific. A generalised loss function is not tailored to a specific problem in isolation, but assesses overall qualities of a generated or predicted instance, such as the difference between pixels in a generated image and those in a ground truth \cite{MSE1, MSE2}. Contrary to this is the development of a specific or custom loss function. This is an involved process requiring knowledge of the important aspects of the optimisation being undertaken and the features of the data to focus on \cite{PRL_GAN_loss}. Tailored loss functions are often problem-dependent, relying on a deep understanding of the domain and specific task at hand. This is challenging when applied to the generation of fluid flow fields, as these specific loss functions \cite{CFD_loss_3}, make comparisons between related works difficult, as the learnt performance of two identical architectures may be different. Additionally, any loss functions utilized for deep learning must be robust enough to measure deviations or convergences concerning a goal, whilst also exhibiting stability in evaluating performance. Loss function instability can significantly impact the effective training of machine learning architectures by not allowing a network to quickly or effectively capture, learn and subsequently represent the underlying data distribution \cite{loss_stability}. Selecting a suitable loss function is a non-trivial decision, often requiring tailoring it to the specific nuances of the problem at hand.
\\ \\
This paper proposes, develops and tests a dynamic loss function that identifies, extracts and uses the gradient containing regions of an image to produce higher quality flow field predictions at faster rates throughout the training process by automatically and dynamically assigning different weights to different regions of the field, based on estimating the local velocity gradient. Additionally, a further variation of the loss function is tested where a priori knowledge is used to further accelerate training speed, loss convergence and image quality by varying known optimal loss function parameters during training. The loss function was implemented in a way that allows for successful operation regardless of the dataset in question, where the weighting is determined on a case by case basis without adaptation or modification for any specific features of a given  dataset. The approach is based on the commonly used Mean Square Error (MSE) loss \cite{MSE1, MSE2}, and is modified to automatically identify and dynamically produce a weighting relating to the regions of greatest pixel disparity according to their gradient intensities. The motivation for the development is discussed (Section \ref{RelatedWork}), with the method detailed and equations provided in full (Section \ref{Method}). The experimental design and algorithm (Section \ref{Experiments}) are provided, with the results (Section \ref{Experiments}) explained, after the proposed method was comprehensively evaluated. Conclusions are drawn in Section \ref{Conclusion}.

\section{Related Work}
\label{RelatedWork}

Deep learning architectures, particularly in generative tasks, rely heavily on loss functions to guide the learning process. These functions quantify the discrepancy between generated data and ground truth, enabling the model to adjust its parameters and improve its output. While generic loss functions like Mean Squared Error (MSE), Cross-Entropy (CE), and Binary Cross-Entropy (BCE) are widely used, their inherent limitations necessitate exploring tailored alternatives, especially for complex domains like Computational Fluid Dynamics (CFD). MSE loss, a common choice for image generation \cite{MSELossOG}, operates by calculating the average squared difference between corresponding pixels in the generated image and the ground truth \cite{MSEloss_image_2}. Its effectiveness stems from its straightforward pixel-by-pixel comparison, assuming equal importance of all pixels \cite{MSEloss_image_1}. However, this assumption can be problematic in applications where specific regions or features hold greater significance.
\\ \\
CE and BCE loss, on the other hand, are frequently employed in classification tasks involving probability distributions \cite{BCELossOG}. CE loss measures the difference between the predicted probability distribution and the actual distribution, making it suitable for multi-class scenarios. BCE loss, a special case of CE, is specifically designed for binary classification, evaluating the difference between the predicted probability of the positive class and the true binary label \cite{BCELossOG}. Both CE and BCE are extensively utilized in training deep learning architectures such as neural networks (NNs), convolutional neural networks (CNNs) and generative adversarial networks (GANs).
\\ \\
GANs have emerged as a powerful tool in fluid dynamics, particularly because of their ability to learn from limited datasets \cite{SmallSetGAN} – a crucial advantage given the cost and difficulty of generating large CFD datasets \cite{CompFluids3}. These networks consist of a generator, which synthesizes new data instances, and a discriminator, which attempts to differentiate between generated and real samples. Loss functions play a critical role in training both networks, providing feedback on the generator's performance and the discriminator's ability to distinguish real from fake data \cite{BCEinGAN}. GANs, in their original and conditional (cGAN) forms, have been successfully applied to various fluid flow problems, including laminar vortex shedding prediction \cite{work1}, airfoil flow field reconstruction \cite{work2}, and supercritical airfoil analysis \cite{work3}. These networks have been trained with a range of loss functions, from basic implementations \cite{CompFluids5} to highly specialized variants \cite{CompFluids4}.
Despite their successes, generic loss functions like MSE, CE, and BCE loss suffer from a key limitation: they treat all errors equally, irrespective of their relative importance within the specific application \cite{lossfunctionsstats}. This equal weighting scheme fails to capture the intricacies of complex data distributions and relationships \cite{MSEloss_image_1}, potentially leading to generated instances that exhibit superficial similarity to the ground truth while missing crucial underlying features \cite{feature_loss}. While modifications to these generic functions have been proposed to address this issue, they often result in highly specialized solutions tailored to a single problem or dataset \cite{PRL_GAN_loss} \cite{PRL_CE_Loss} \cite{CFD_LossFunction}. These specialized functions lack generalizability and require substantial re-engineering for adaptation to different datasets, making cross-study comparisons challenging.
\\ \\
Consequently, a pressing need exists for a loss function that is general in nature, but can identify, weight and score the regions of importance in fluid field generation. Such a loss function needs to be implemented such that it does not require complex modification or manual intervention by researchers \cite{GeneralLoss} when applied to different contexts or datasets. To achieve this, a novel adaptation to the MSE loss function is proposed. This modification seeks to automatically and dynamically identify the regions where the field changes quickly. This gradient region is then used to weight the loss between the generated field and a ground truth CFD dataset instance on a field by field basis. The determination of gradient weighting is designed such that it will work on any given velocity, pressure and density field, to ensure broad applicability. The novel loss function is further explained in Section \ref{Method}.

\section{Method} 
\label{Method}

The most widely applicable loss function for general image reconstruction, the Mean Squared Error (MSE) loss \cite{MSE1, MSE2}, is used to determine a scalar (0D) mean error value between a ground truth instance ($I_R$) and a secondary, or generated instance ($I_G$) by subtraction and then summation of the error in each direction ($h, w$) over $n$ instances. The MSE loss is denoted by Eq. (\ref{Normal_MSE_eqn}). The proposed novel method differs from Eq. (\ref{Normal_MSE_eqn}) by seeking to dynamically identify the regions of importance in the flow fields where strong gradients of interest are present and then weight these regions accordingly in the loss calculation. Computational fluid dynamics data is well suited to this, as strong gradients hold importance in the evaluation of the underlying data. To achieve this custom loss, the MSE loss, Eq. (\ref{Normal_MSE_eqn}), has been modified to include a 2D dynamic weighting ($W_i$) for each generated image ($I_G$), where $W_i$ is produced from the disparity of the known CFD ground truth instance ($I_R$) on an field-by-field basis. The detailed process to determine this weighting is provided in Section \ref{The_Loss_Function}. The modified form of the MSE loss equation is given below, Eq. (\ref{GMSE_eqn}), and it is called Gradient Mean Squared Error loss (GMSE).

        \begin{equation}
        \label{Normal_MSE_eqn}
        MSE = \frac{1}{n} \sum_{i=1}^{n} \left[ \frac{1}{h w} \sum_{j=1}^{h} \sum_{k=1}^{w}\left ( I_{R}(j,k)-\hat{I_{G}(j,k)} \right )^2 \right]
        \end{equation}

        \begin{equation}
        \label{GMSE_eqn}
        GMSE = \frac{1}{n} \sum_{i=1}^{n} \left[ \frac{1}{h w} \sum_{j=1}^{h} \sum_{k=1}^{w} W_{i}\left  ( I_{R}(j,k)-\hat{I_{G}(j,k)} \right )^2 \right]
        \end{equation}

The following sections detail the conceptualisation, development and testing of the dynamic weighting function. The assessment of it, the test architecture, implementation and algorithm structure are provided, as are the experimental specifics.

\subsection{Computational fluid dynamics data} 
\label{CFD_Data_explanation}

\begin{figure}[H]
        \centering
        \includegraphics[scale=.75]{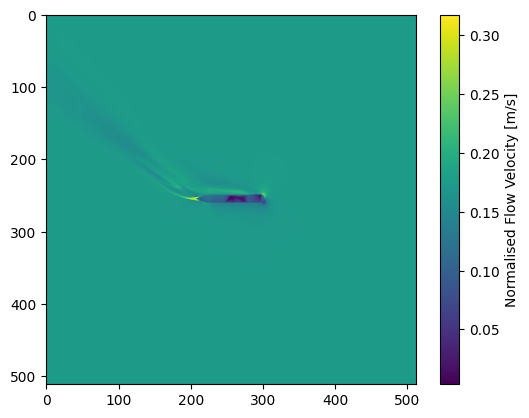}
        \caption{Ground truth CFD plane, used to determine pixel weights. Central body geometry of a submarine is shown in the centre, as observed from a top down perspective. Higher flow velocity indicated by yellow gradient regions.}
        \label{fig:GroundTruth}
        \end{figure}

CFD data is intrinsically highly dynamic, gradient driven data. CFD data is produced in the assessment of fluid flow around, or through, an object of interest \cite{cfd_data_uses}. When the fluid moves its velocity, pressure and density change in difficult to predict ways, due to the turbulent nature of fluid flow. Subsequently, the results of such analyses contain small regions of interest, with highly varying gradients. The cross sections of these regions are often represented in planar, 2D formats, similar to traditional 2D images \cite{cfd_data_uses}, making them an ideal candidate for modified generative ML frameworks. An example cross section of such a CFD simulation is provided in Fig. \ref{fig:GroundTruth}, showing a central submarine body, with a fluid flow field (velocity magnitude) around it on a 2D plane. This image shows a highly complex separated flow structure, that is of interest to fluid dynamics researchers. The structure, driven by the submarines propeller, exhibits a large degree of separation from the body and erratic flow structure behaviour due to turbulence. The variant regions, of this wake structure, whilst small in size, are important in the subsequent analysis of the flow field, as they can impact the vessel's drag and lift characteristics. Optimising the generation and accurate approximation of flow fields such as this and their structures, through deep learning is a primary focus of current research \cite{CFD_GAN_Buildings}. The next sections will detail the process to identify and dynamically weight these regions using the proposed GMSE loss function. 

\subsection{Network Architecture} 
\label{GAN_Architecture}

\begin{figure}[H]
        \centering
        \includegraphics[scale=0.25]{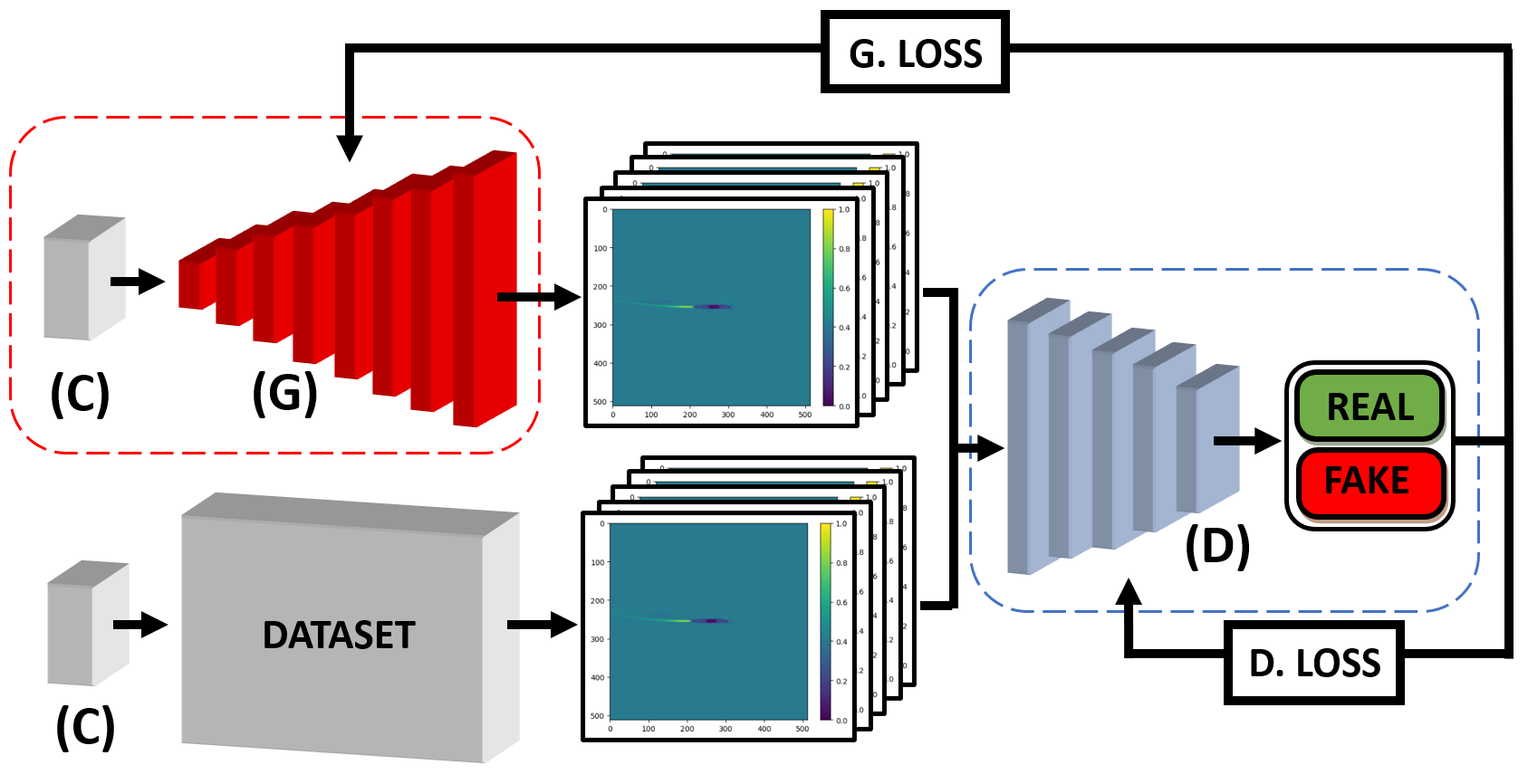}
        \caption{Schematic diagram of the controlled generative adversarial (CGAN) network.}
        \label{fig:CGAN_structure}
        \end{figure}

A neural network is required to test the effectiveness of the proposed GMSE loss function. A controlled generational adversarial network (CGAN) was selected to serve this purpose. The CGAN architecture was comprised of two neural networks that worked against each other. A schematic diagram of this architecture is provided in Fig. \ref{fig:CGAN_structure}. The first neural network, the Generator (G), was responsible for generating 512x512 grey-scale images of the flow fields, when given the speed and angle of the vessel. A discriminator (D) was used to view both the generated field and a ground truth field from the CFD dataset and determine if the generated field was a true dataset instance or not. The entire architecture was controlled (C) so that the generator was constrained to generate a field based on the speed and angle data it was passed. The architecture of the CGAN was be held constant across testing, with only the loss function used varied to assess the image quality with different modifications to the loss function. Further detail is provided in Section \ref{Related_Methods}.

The explicit architecture of the generator and discriminator are provided in Appendix \ref{app1}. The generator architecture took a controlled 1-dimensional latent space in the form of a vector with shape (1, 128) that specified the parameters (flow speed, heading and direction) that the generated image should contain and reshaped it into a 4D tensor, for input into the first convolutional transpose layer. A series of linearly increasing transposed convolutions were used to gradually increase the spatial dimensions whilst learning features \cite{progressiveGAN}. The spatial resolution was increased and up-sampling undertaken by a factor of 2, allowing for a controlled increase in each layer \cite{Conv2dTranpose}. This occurred in layers 1 through 7. A mixture of ReLU and Leaky ReLU activation functions were used to mitigate the vanishing gradient problem in the deeper layers of the architecture \cite{NeuronDecay}. Batch normalisation was used to normalise the later weights to reduce internal co-variate shift \cite{BatchNorm}. The generator network was optimised using either the MSE loss or the GMSE loss value, which was then back-propagated through the architecture for optimisation.
\\ \\
The discriminator architecture based on a CNN structure, utilizing 4 convolutional layers were used to take the single channel generated image, extract its features, and then feed it into a fully connected linear layer after flattening. Convolutional layers were used to allow the discriminator to learn from features at different spatial scales whilst the dimensionality of the data was reduced \cite{StridedConvols}. Leaky ReLU was again used to address vanishing gradients in the deeper layers and batch normalisation used to stabilise discriminator training \cite{NeuronDecay}, \cite{BatchNorm}. The fully connected linear layer then took the feature vector and mapped it to a single output value, to which the Sigmoid activation function was applied, resulting in a classification between 0 (fake) and 1 (real) \cite{Sigmoid}. This output represented the probability that the image was real (belonging to the CFD dataset, and not the generator). The discriminator was then optimised using cross entropy loss. 

\subsection{Gradient Mean Squared Error (GMSE) Loss} 
\label{The_Loss_Function}

\begin{figure}[!ht]
        \centering
        \includegraphics[scale=0.3]{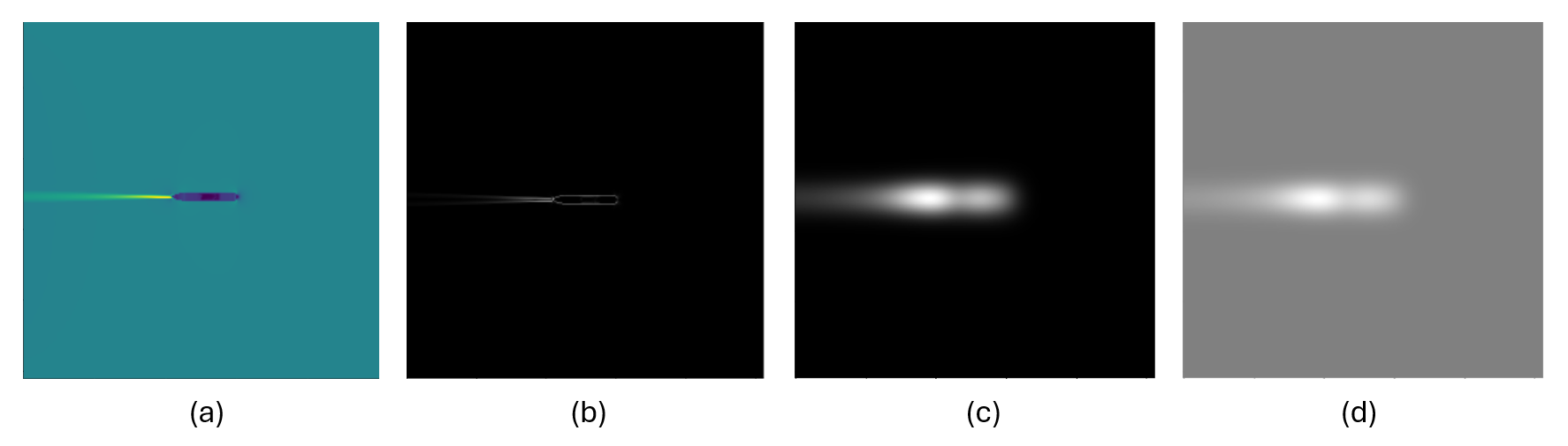}
        \caption{Visual depiction of GMSE loss function operation. The reference ground truth CFD image from the dataset (a) is used to first produce a disparity array (b). The disparity array is then blurred, resulting in the blurred array (c). Finally, a weighting array (d) is produced, used to score the individual pixel loss during training for each new instance the generator provides. The grey appearance of (d) results due to the non-zero lowerbound, used to give weighting to the freestream of the flowfield.}
        \label{fig:VisualGMSE}
        \end{figure}

The Gradient Mean Squared Error (GMSE) is designed to operate using the change in the field to generate more accurate predictions. The operation of the GMSE loss function is sufficiently generalised to allow for velocity, pressure or density fields to be used to determine the weighting used for the loss calculation. To calculate the Gradient Mean Squared Error (GMSE) loss, a series of modifications were made to the MSE loss equation. A visual depiction of this process is provided in Fig. \ref{fig:VisualGMSE}, showing the process undertaken to determine the 2D weighting array (Fig. \ref{fig:VisualGMSE}d) for the GMSE loss function operation using the ground truth flow field (Fig. \ref{fig:VisualGMSE}a), disparity array (Fig. \ref{fig:VisualGMSE}b) and blurred array (Fig. \ref{fig:VisualGMSE}c). 

First, element-wise subtraction was conducted with respect to the x (Eq. \ref{pixel_eqn1}) and y axis (Eq. \ref{pixel_eqn2}) of the ground truth field. This determined the gradient regions by the relative difference between adjacent cells/pixels, similar to a unidirectional spatial highpass filter. $W_{d,x}$ denotes the x-axis pixel disparity. This is comprised of the amplitude of each pixel $W_x$, subtracted by the amplitude of the previous pixel $W_{x-1}$. This was then also conducted for the y axis, $W_{d,y}$ using $W_{y}$ and $W_{y-1}$. $W_{d}$ was then calculated as the magnitude of the resulting vector addition of $W_{d,x}$ and $W_{d,y}$ (Eq. \ref{pixel_eqn3}).

    \begin{equation}
        \label{pixel_eqn1}
        W_{d,x} =  W_{x} - W_{x-1}    
        \end{equation}

    \begin{equation}
        \label{pixel_eqn2}
        W_{d,y} =  W_{y} - W_{y-1} 
        \end{equation}

    \begin{equation}
        \label{pixel_eqn3}
        W_{d} = \sqrt{W_{d,x}^{2} + W_{d,y}^{2}}
        \end{equation}

\begin{figure}[!ht]
        \centering
        
        \subfloat[]{\includegraphics[scale=0.50]{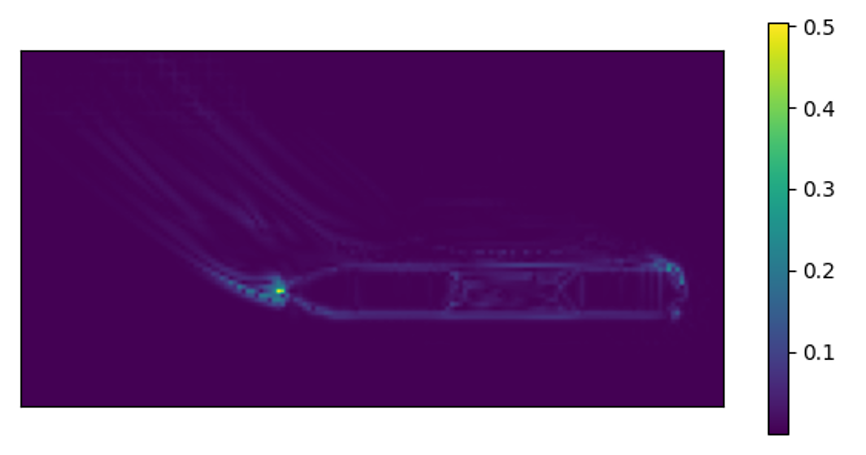}
        \label{fig:Disparity_1}}\\ 
        
        \subfloat[]{\includegraphics[scale=0.50]{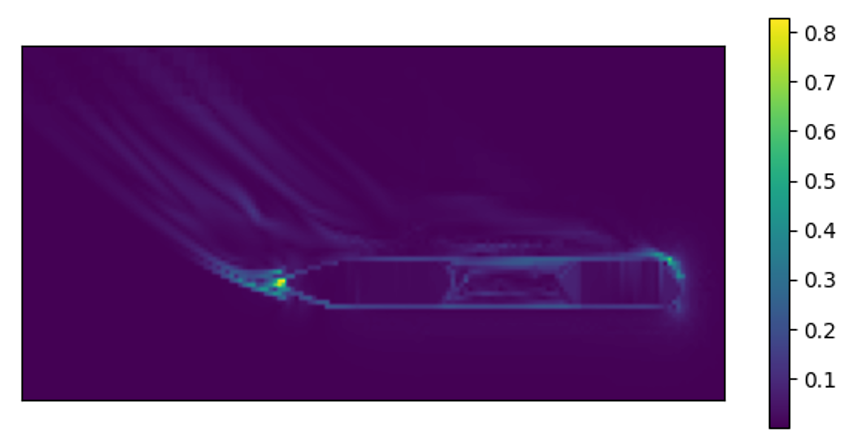}
        \label{fig:Disparity_2}}
    
        \caption{Image is a cropped version of the flow field depicted in Fig. \ref{fig:GroundTruth}. Detailed is the pixel disparity using a standard difference of Gaussians (DOG) filter (a) and the non-linear disparity determination from this method (b). Absolute disparity magnitude is indicated by the colorbars, as shown. Additional detail is shown regarding the disparity on (b) when compared to (a)}
        \label{fig:Disparity}
    \end{figure}

This operation (\ref{pixel_eqn3}) is non-linear. The disparity calculation was undertaken to ensure that small scale disparities were included in the disparity array. Traditional difference of Gaussians (DOG) filters with linear addition, unlike Eq. (\ref{pixel_eqn3}), can produce reduced disparity where inverse signs exist. The modification for non-linear addition, Eq. (\ref{pixel_eqn3}), was found to result in increased disparity capture, particularly in high frequency but low amplitude regions of interest. The difference between this operation and a normal DOG filter, followed by rectification, can be seen in Fig. \ref{fig:Disparity}. Provided in Fig. \ref{fig:Disparity}(a) is the result of a typical DOG filter. The non-linear disparity operation is provided in Fig. \ref{fig:Disparity}(b).

A Gaussian blur was then applied to the disparity array to blur and increase the size of the areas of importance, where the disparity occurred. The Gaussian function is given by Eq. (\ref{Gaussian_eqn}), where $x$ is the distance from the origin in the lateral direction and $y$ is the distance from the origin in the vertical direction. In Eq. (\ref{Gaussian_eqn}) the $\sigma$ value, denotes the standard deviation of the Gaussian distribution, which can be manipulated to shift the values during the blur process, depending on the desired extent of the salience expansion from the point of variation.

\begin{equation}
        \label{Gaussian_eqn}
        W_{blur}(x,y)=\frac{1}{2 \pi \sigma^2}e^{-\frac{x^2 + y^2}{2 \sigma^2}}
        \end{equation}

Next, the blurred array ($W_{blur}$), was raised to the power of a fixed value, denoted by gamma ($\gamma$), to strengthen or weaken the gradient of the blur in the resulting array ($W_{\gamma}$). This modification resulted in different weighting values and was be used to affect the final weighting and performance of the loss function. The formula for this is given in Eq. (\ref{Gamma_eqn}). 

\begin{equation}
        \label{Gamma_eqn}
        W_{\gamma} = W_{blur}^{\gamma}
        \end{equation}

The modified array ($W_{\gamma}$) was then normalised between its maximum and minimum values in accordance with Eq. (\ref{Normalisation_eqn}). Thus all values, for the normalised weight, $W_{norm}$, fell in the range [0,1]. A $W_{norm}$ lower bound of 0 can have a negative impact on image generation, where it would indicate that these regions of the image are unimportant. This is not the case, as the region away from the highest disparity areas (the freestream in this case) still retains a limited relative importance. To ensure this is accounted for $W_{norm}$ is adjusted by an offset to achieve a non-zero lower bound. 

\begin{equation}
        \label{Normalisation_eqn}
        W_{norm} = \frac{W_{\gamma} - min(W_{\gamma})}{max(W_{\gamma}) - min(W_{\gamma})}
        \end{equation}
   
    \begin{equation}
        \label{Offset_eqn}
        W_{i} = \left (W_{norm} \cdot \left [ 1 - C_{o} \right ]  \right ) + \left ( C_o\right )
        \end{equation}

The offset $C_o$ was applied, resulting in the range [$C_o$, 1]. Larger values of $C_o$ increased the relative weight of the surrounding gradient region compared to the strongest gradient structures. The final weighting, developed from the ground truth image, was then determined and is given by Eq. (\ref{Offset_eqn}). This weighting then applied within the GMSE loss function as given by Eq. (\ref{GMSE_eqn}) to yield the GMSE weighting for each set of image. This loss is used to score the quality of the generated instances produced at each epoch of training. 

\subsection{Algorithm} \label{GMSE_Algorithm}
Algorithm 1 shows the implementation of the equations required to determine the gradient mean squared error loss between a set of generated images (F) and a set of real images (R). To begin, F and R were both moved to the GPU, in the interest of compute speed. Next, a disparity tensor (D) was constructed. After which on an image-by-image basis using R, the disparity (x, y) was determined, negated and normalised, before being added to D. Then a Gaussian blur was applied to all images within D, in accordance with the selected sigma value. Then a gamma operation was performed, before the blurred arrays were normalised between 0 and 1, and offset used to include a relative importance for all regions of the image. After which the dynamic weighted squared error was computed and then the mean of these values taken. This GMSE value across all images was then returned and used for loss propagation through the generator network.

\begin{algorithm}[H]
    \label{GMSEAlgo}
    \caption{Gradient Mean Squared Error Loss}\label{alg:GMSE}
    \begin{algorithmic}
    \STATE \textbf{Input:} Fake images $\mathbf{F}$, Real images $\mathbf{R}$, $\gamma$, $\sigma$, $C_{o}$
    \STATE \textbf{Output:} GMSE loss
    \STATE 
    \STATE Initialize disparities tensor $\mathbf{D} \gets \emptyset$
    \FOR{each real image $\mathbf{r} \in \mathbf{R}$}
        \STATE $\mathbf{d} \gets \text{diff}(\mathbf{x}) \& \text{diff}(\mathbf{y})$ \COMMENT{Calculate disparity (x, y)}
        \STATE $\mathbf{d} \gets \sqrt{\mathbf{x}^2 + \mathbf{y}^2}$ \COMMENT{Calculate magnitude}
        \STATE Normalize $\mathbf{d}$ to $[0, 1]$ \COMMENT{Not required for $\gamma = 1.00$}
        \STATE $\mathbf{D} \gets \text{append}(\mathbf{D}, \mathbf{d})$
    \ENDFOR
    \STATE Apply Gaussian blur to $\mathbf{D}$ with $\sigma$ and kernel $ = \sigma \cdot 6 + 1$
    \STATE $\mathbf{W} \gets \mathbf{W}^\gamma$ \COMMENT{Gamma operation}
    \STATE Normalize $\mathbf{W}$ to $[0, 1]$ using min-max normalization
    \STATE Apply offset: $\mathbf{W} \gets (1-C_{o})\mathbf{W} + C_{o}$
    \STATE Calculate difference: $\mathbf{DSE} \gets \mathbf{W} * (\mathbf{R} - \mathbf{F})^2$
    \STATE Calculate mean difference: $\text{GMSE} \gets \text{mean}(\mathbf{Diff})$
    \STATE \textbf{return} $\text{GMSE}$
    \end{algorithmic}
\end{algorithm}

\section{Experiments} \label{Experiments}
This section provides a detailed overview of the dataset utilised for the experiments and the testing methodology. The evaluation metrics employed to assess the impact of parameter variation on the proposed loss metric are outlined, along with the settings and results derived from the conducted experiments.

    \subsection{Dataset} \label{Dataset}

    A dataset of submarine flow was used. This dataset contained 1200 distinct instances. In each instance, a different flow speed and flow angle was used, and the movement of fluid around a submarine was determined. Each instance is a fully resolved computational fluid dynamics (CFD) assessment, using the k-epislon turbulence model in conjunction with the Reynolds Averaged Navier Stokes (RANS) equations as outlined in an existing work \cite{ZacIEEE}. Each CFD assessment in the dataset was then interpolated onto a uniform 2D cross sectional plane (as in Fig. \ref{fig:GroundTruth}) containing the velocity magnitude of the flow. The speed of the flow hitting the submarine was varied from 0.1 m/s to 5.0 m/s and the angle of impact varied from 0 degrees to 60 degrees. This dataset was observed to contains small regions of interest where strong gradients were present, making it ideal to assess the custom GMSE loss function. Each instance in the dataset was developed using ANSYS Fluent, a commercial computational fluid dynamics software. Further details on the generation of this dataset can be found in \cite{ZacIEEE}.

    \subsection{Comparison with related methods} \label{Related_Methods}

    The performance of the novel loss function was assessed via a comparison to the existing Mean Squared Error (MSE) loss, as well as two baseline variants of the GMSE loss function. The first variant (GMSE) had a fixed set of parameters ($\sigma, \gamma, C_{o}$) and the second (DGMSE) had them dynamically changed during training. The GMSE loss function had the parameters fixed such that $\sigma = 10, \gamma = 1.00$ and $C_{o} = 0.2$. DGMSE was used to assess the effect of beginning with a high blur strength (high $\sigma$) and a focus on the gradient containing region specifically (low $C_{o}$), before switching to a more equal treatment of the image. The DGMSE loss function had (30, 0.20, 0.1) for the first epoch, then (20, 0.40, 0.1) for epochs 1 to 5, before (20, 0.40, 0.2) was adopted until 20 epochs and then (25, 0.40, 0.2) fixed until the end of training. These settings were selected based on analysis of the results from an exploration of the effect of different parameters on network training performance.
\\ \\  
    The architecture used to provide the assessment of the loss functions was fixed. The generator of the controlled generational adversarial network (cGAN) from Fig. \ref{fig:CGAN_structure} was optimised using the different loss functions one at a time. Each loss function was used to optimise the generator of the CGAN architecture for a duration of 300 epochs, until convergence was reached. The hyper-parameters used in this training loop were held constant across all tests. During training, the generator loss value, discriminator loss value and the structural similarity index measure (SSIM) were recorded. The justification and explanation of these metrics are provided in Section \ref{Evaluation_Metrics}.  
\\ \\   
    After the baselines had been tested, variations of the GMSE loss equations were used where the sigma ($\sigma$), gamma ($\gamma$) and offset ($C_{o}$) values were incremented to assess the effect of non-linear shifts, blur strength and relative weighting respectively. The gamma value was incremented between 0.20 and 2.00 in 0.20 increments. Sigma was varied and tested at $\sigma = 2, 5, 10, 20, 30$ and $40$. The offset values were varied between 0.1 and 0.9 in 0.1 increments. To ensure the testing was comparable, fixed seeds were used to set the random NumPy variables used in testing and the Torch seed. This ensured that all generative activities, optimisation and back-propagation would begin from the same initial start-point. 

\subsection{Evaluation metrics} \label{Evaluation_Metrics}

    The accuracy and convergence of the proposed method must be evaluated with respect to a known ground truth. Additionally, the evaluation of the accuracy needs to be sufficiently complex to capture the spatial and flow driven relationships present in the data. To ensure this, the effectiveness of the generative network and the loss functions themselves were monitored in several ways. 
\\ \\
    To make an analysis of the loss convergence during training, the loss was recorded at both the generator and discriminator for both baseline optimised networks as well as all GMSE variants. During training of the generator, the generator loss was recorded using the validation set, for each batch, within each epoch. From this, the batch averaged generator loss at each epoch was produced. This approach was replicated to record the discriminator loss for each batch, within each epoch, resulting in the batch averaged discriminator loss at each epoch. These loss curves were then normalised to a maximum value of 1. The normalised rate of loss per epoch ($\frac{\mathrm{d} }{\mathrm{d} \epsilon } L_{N,max}$) is also recorded, to provide comparison with regard to the rate of convergence.
\\ \\
    To assess the quality of the generated fields as the network learnt, the network was progressively saved over epochs. Both the generator architecture and its associated weights were saved at epochs 1, 5, 10, 20, 50, 100 and 300. Whilst generated fields can have similar MSE values, but different qualitative performance \cite{SSIMvsMSE}, the structural similarity index measure (SSIM) was also determined at 1, 5, 20, 100 and 300 epochs. The SSIM gives a better indication of the structural similarity between the generated flow field and the ground truth RANS flow field. The SSIM determines the image degradation where there are changes, or perceived changes to the structural information of a field \cite{SSIM} occur. The structural similarity index was computed by application of Eq. \ref{SSIMeqn}. Where the two fields (generated and RANS) are used, as denoted by (x, y). The pixel mean is given by $\mu$ and the variance $\sigma$, specific to each field and the covariance by $\sigma_{x,y}$. Two constants ($c_1, c_2$) were used to stabilise the division operations \cite{SSIM_2}. The two constants are given by Eq. \ref{c1eqn} and \ref{c2eqn} respectively, where $L$ denotes the dynamic range of the pixel-values, $k_{1}$ and $k_{2}$ are fixed constants of 0.01 and 0.03 respectively, as per \cite{SSIM_constants}. 

    \begin{equation}
        \label{SSIMeqn}
        SSIM(x,y) = \frac{(2 \mu_{x} \mu_{y} + c_1)(2 \sigma_{x,y}+c_2)}{(\mu_{x}^{2} + \mu_{y}^{2} + c_1)(\sigma_{x}^{2} + \sigma_{y}^{2} + c_2)}
    \end{equation}

    \begin{equation}
        \label{c1eqn}
        c_{1} = (k_{1} \cdot L)^{2}
    \end{equation}

    \begin{equation}
        \label{c2eqn}
        c_{2} = (k_{2} \cdot L)^{2}
    \end{equation}
    
    In addition to the quantitative metrics, a set of fields were produced for qualitative comparison between the MSE and GMSE baseline optimised networks during training for a number of different epochs. 

    \subsection{Experimental setting} \label{Experimental_Setting}

    Experimental testing was conducting using a PC with an AMD Threadripper Pro 5955WX, 256GB of RAM and 2 NVIDIA A6000 GPUs. The testing was conducted using Python 3.11.6 and the Pytorch (2.0.0 + cu117), Numpy (1.22.4), Pandas (2.1.4) and Kornia (0.7.1) libraries. To assess the loss functions, the networks were optimised for 100 epochs, after which GMSE convergence was seen. The learning rate of the generator architecture was fixed to $2 \times 10^{-5}$ and $1.5 \times 10^{-5}$ for the discriminator. A batch size ($n$) of 20 images was used with the Adam optimizer used for both generator and discriminator.

\section{Results} \label{Results}

    Table \ref{tab:data_table} presents the structural similarity index measure (SSIM) at epochs 1, 5, 20, 100 and 300. The maximum normalised rate of loss is also indicated. The standard deviation ($\sigma_{s.d.}$) is also provided to give an indication of the performance variability with respect to modifying the gamma ($\gamma$), sigma ($\sigma$) or offset ($C_{o}$) parameters of the baseline GMSE loss function. The MSE loss optimised network performance and that of the dynamic gradient mean squared error loss function (DGMSE) are also provided. Indicated in bold is the GMSE baseline for the gamma ($\gamma$), sigma ($\sigma$) and offset ($C_{o}$) variations. 
\\ \\
    The GMSE and DGMSE loss functions were seen to outperform the MSE loss function in both SSIM and maximum loss rate across all epochs of training. At early epochs, when the SSIM is low, the GMSE (0.068) and DGMSE (0.075) still outperformed MSE optimisation (0.032). The disparity between the approaches widened further through training, where the GMSE and DGMSE loss functions proved beneficial, reporting final SSIM's of 0.988 and 0.989 respectively, compared to 0.933 for the MSE loss function. Marking a reduction in structural dissimilarity of 82.1\% and 83.6\% in comparison to the MSE loss. An increased rate of maximum loss was also observed for the GMSE and DGMSE variants of -0.143 (33.6\% higher) and -0.189 (76.6\% higher) respectively, compared to -0.107 for the MSE loss function. This stronger rate of normalised loss convergence indicates that the network can better optimise in response to the initial loss between the generated instances and the real dataset instances. 
\\ \\
    Variations to the gamma ($\gamma$) variable were seen to influence both the SSIM and the rate of loss. The standard deviation of gamma variation was observed to be larger (0.082, 0.108 and 0.146) during early training (1, 5 and 20 epochs) before reducing to smaller values (0.033, 0.018) towards the completion of training (100 and 300 epochs). This correlates to the close SSIM observed at later training stages, with most structural similarities being clustered around 0.983 to 0.989, with a maximum recorded SSIM of 0.991. The maximum SSIM outperformed the MSE baseline by 86.6\%. There is more variance reported in the rate of loss recorded for the gamma variations, with a standard deviation of 0.017, a maximum of -0.177 ($\gamma = 0.60$) and a minimum of -0.115 ($\gamma = 2.00$). $\gamma > 1.20$ are seen to correspond to reduced rates of loss, and variations where $\gamma < 1.20$ correspond to an increased rate of loss. 
\\ \\
    Variations to sigma ($\sigma$) were also observed to effect the SSIM and the rate of loss, with different performance depending on the strength of the blur. Higher blur strengths ($20 < \sigma < 40$) were seen to correspond the highest SSIM at epochs 1 (0.285), 5 (0.654), 20 (0.926) and 100 (0.982). Blur strengths between 10 and 30 also corresponded to the highest rates of loss (-0.143, -0.141 and -0.133), outperforming the MSE baseline by 33.6\%, 31.7\% and 24.3\% respectively. Again, the variation of SSIM was seen to decrease as training progressed, with a standard deviation of 0.022 reported at the completion of training. 
 \\ \\   
    Variations to the offset ($C_{o}$) had the largest SSIM deviation at the completion of training (0.028) indicating that the offset weighting can still affect the image quality further into the training cycle. This is reflected by the higher SSIM values recorded for offsets of 0.1, 0.2 and 0.3, which highly weight the gradient regions. These offset values corresponded to the highest SSIM values recorded for the offsets (0.986, 0.988, 0.976 respectively). The opposite is true of $C_{o} > 0.3$ which were observed to record SSIMs between 0.923 and 0.926, thus under performing the MSE loss by 10.5\% to 14.9\%. It should be noted that all variations to the baseline and offset (($C_{o}$) still reported higher rates of loss (-0.110 to -0.158) than the MSE baseline by a minimum of 2.8\% to a maximum of 47.7\%.

    \begin{table}[H]
    \caption{Structural similarity index (SSIM) and normalised generator loss rates for the baseline MSE and GMSE optimised networks with variations to gamma ($\gamma$) variants, sigma ($\sigma$) and the offset ($C_{o}$). Bolded baseline GMSE indicated in each respective parameter variation.}
    \label{tab:data_table}
    \setlength{\tabcolsep}{3.5pt}
    \begin{tabular}{|ll|lllll|l|}
    \hline
    \multirow{2}{*}{\textbf{Variant}} & \multirow{2}{*}{\textbf{\#}} & \multicolumn{5}{l|}{\textbf{SSIM at epoch}} & \multirow{2}{*}{\textbf{$\frac{\mathrm{d} }{\mathrm{d} \epsilon } L_{max}$}} \\ \cline{3-7}
                             &  & \multicolumn{1}{l|}{\textbf{1}}  & \multicolumn{1}{l|}{\textbf{5}}  & \multicolumn{1}{l|}{\textbf{20}}  & \multicolumn{1}{l|}{\textbf{100}}  & \textbf{300}  &   \\ \hline
    Baselines                      & MSE & \multicolumn{1}{l|}{0.032} & \multicolumn{1}{l|}{0.239} & \multicolumn{1}{l|}{0.456} & \multicolumn{1}{l|}{0.829} & 0.933 & -0.107                    \\ 
                         & \textbf{GMSE} & \multicolumn{1}{l|}{\textbf{0.068}} & \multicolumn{1}{l|}{\textbf{0.617}} & \multicolumn{1}{l|}{\textbf{0.833}} & \multicolumn{1}{l|}{\textbf{0.972}} & \textbf{0.988} & \textbf{-0.143} \\
                         & DGMSE & \multicolumn{1}{l|}{0.075} & \multicolumn{1}{l|}{0.674} & \multicolumn{1}{l|}{0.935} & \multicolumn{1}{l|}{0.978} & 0.989 & -0.189\\ \hline
    
    \multirow{10}{*}{$\gamma$}  & 0.20 & \multicolumn{1}{l|}{0.282} & \multicolumn{1}{l|}{0.474} & \multicolumn{1}{l|}{0.918} & \multicolumn{1}{l|}{0.978} & 0.991 & -0.124                    \\  
                             & 0.40 & \multicolumn{1}{l|}{0.167} & \multicolumn{1}{l|}{0.632} & \multicolumn{1}{l|}{0.940} & \multicolumn{1}{l|}{0.983} & 0.990 & -0.142                    \\ 
                             & 0.60 & \multicolumn{1}{l|}{0.185} & \multicolumn{1}{l|}{0.557} & \multicolumn{1}{l|}{0.916} & \multicolumn{1}{l|}{0.980} & 0.989 & -0.177                    \\  
                             & 0.80 & \multicolumn{1}{l|}{0.103} & \multicolumn{1}{l|}{0.593} & \multicolumn{1}{l|}{0.847} & \multicolumn{1}{l|}{0.974} & 0.984 & -0.130                    \\  
                             & \textbf{1.00} & \multicolumn{1}{l|}{\textbf{0.068}} & \multicolumn{1}{l|}{\textbf{0.617}} & \multicolumn{1}{l|}{\textbf{0.833}} & \multicolumn{1}{l|}{\textbf{0.972}} & \textbf{0.988} & \textbf{-0.143}                    \\
                             & 1.20 & \multicolumn{1}{l|}{0.053} & \multicolumn{1}{l|}{0.561} & \multicolumn{1}{l|}{0.722} & \multicolumn{1}{l|}{0.961} & 0.983 & -0.120                    \\ 
                             & 1.40 & \multicolumn{1}{l|}{0.038} & \multicolumn{1}{l|}{0.487} & \multicolumn{1}{l|}{0.715} & \multicolumn{1}{l|}{0.949} & 0.983 & -0.121                    \\  
                             & 1.60 & \multicolumn{1}{l|}{0.026} & \multicolumn{1}{l|}{0.441} & \multicolumn{1}{l|}{0.642} & \multicolumn{1}{l|}{0.936} & 0.976 & -0.130                    \\ 
                             & 1.80 & \multicolumn{1}{l|}{0.036} & \multicolumn{1}{l|}{0.332} & \multicolumn{1}{l|}{0.545} & \multicolumn{1}{l|}{0.905} & 0.954 & -0.121                    \\  
                             & 2.00 & \multicolumn{1}{l|}{0.028} & \multicolumn{1}{l|}{0.307} & \multicolumn{1}{l|}{0.523} & \multicolumn{1}{l|}{0.881} & 0.935 & -0.115  \\ \cline{2-8}
                             
                             & $\sigma_{s.d.}$ & \multicolumn{1}{l|}{0.082} & \multicolumn{1}{l|}{0.108} & \multicolumn{1}{l|}{0.146} & \multicolumn{1}{l|}{0.033} & 0.018 & 0.017 \\ 
                             \hline
                             
    \multirow{6}{*}{$\sigma$}   & 2 & \multicolumn{1}{l|}{0.021} & \multicolumn{1}{l|}{0.281} & \multicolumn{1}{l|}{0.408} & \multicolumn{1}{l|}{0.863} & 0.929 & -0.118                    \\ 
                             & 5 & \multicolumn{1}{l|}{0.040} & \multicolumn{1}{l|}{0.460} & \multicolumn{1}{l|}{0.664} & \multicolumn{1}{l|}{0.943} & 0.981 & -0.126                    \\  
                                 & \textbf{10} & \multicolumn{1}{l|}{\textbf{0.068}} & \multicolumn{1}{l|}{\textbf{0.617}} & \multicolumn{1}{l|}{\textbf{0.833}} & \multicolumn{1}{l|}{\textbf{0.972}} & \textbf{0.988} & \textbf{-0.143}                   \\  
                             & 20 & \multicolumn{1}{l|}{0.213} & \multicolumn{1}{l|}{0.654} & \multicolumn{1}{l|}{0.926} & \multicolumn{1}{l|}{0.982} & 0.989 & -0.141                    \\  
                             & 30 & \multicolumn{1}{l|}{0.285} & \multicolumn{1}{l|}{0.630} & \multicolumn{1}{l|}{0.922} & \multicolumn{1}{l|}{0.980} & 0.991 & -0.133                    \\ 
                             & 40 & \multicolumn{1}{l|}{0.196} & \multicolumn{1}{l|}{0.623} & \multicolumn{1}{l|}{0.915} & \multicolumn{1}{l|}{0.977} & 0.990 & -0.126                    \\ \cline{2-8}
                             
                             & $\sigma_{s.d.}$ & \multicolumn{1}{l|}{0.099} & \multicolumn{1}{l|}{0.134} & \multicolumn{1}{l|}{0.189} & \multicolumn{1}{l|}{0.042} & 0.022 &  0.009 \\ 
                             \hline
                             
    \multirow{9}{*}{$C_{o}$} & 0.1 & \multicolumn{1}{l|}{0.103} & \multicolumn{1}{l|}{0.538} & \multicolumn{1}{l|}{0.923} & \multicolumn{1}{l|}{0.979} & 0.986 & -0.158    \\
                             & \textbf{0.2} & \multicolumn{1}{l|}{\textbf{0.068}} & \multicolumn{1}{l|}{\textbf{0.617}} & \multicolumn{1}{l|}{\textbf{0.833}} & \multicolumn{1}{l|}{\textbf{0.972}} & \textbf{0.988} & \textbf{-0.143}    \\
                             & 0.3 & \multicolumn{1}{l|}{0.036} & \multicolumn{1}{l|}{0.429} & \multicolumn{1}{l|}{0.524} & \multicolumn{1}{l|}{0.949} & 0.976 & -0.126    \\
                             & 0.4 & \multicolumn{1}{l|}{0.020} & \multicolumn{1}{l|}{0.274} & \multicolumn{1}{l|}{0.361} & \multicolumn{1}{l|}{0.859} & 0.924 & -0.134    \\
                             & 0.5 & \multicolumn{1}{l|}{0.019} & \multicolumn{1}{l|}{0.261} & \multicolumn{1}{l|}{0.447} & \multicolumn{1}{l|}{0.874} & 0.924 & -0.133    \\
                             & 0.6 & \multicolumn{1}{l|}{0.017} & \multicolumn{1}{l|}{0.203} & \multicolumn{1}{l|}{0.372} & \multicolumn{1}{l|}{0.875} & 0.926 & -0.132    \\
                             & 0.7 & \multicolumn{1}{l|}{0.025} & \multicolumn{1}{l|}{0.250} & \multicolumn{1}{l|}{0.380} & \multicolumn{1}{l|}{0.855} & 0.926 & -0.115    \\
                             & 0.8 & \multicolumn{1}{l|}{0.024} & \multicolumn{1}{l|}{0.247} & \multicolumn{1}{l|}{0.421} & \multicolumn{1}{l|}{0.841} & 0.923 & -0.118    \\
                             & 0.9 & \multicolumn{1}{l|}{0.027} & \multicolumn{1}{l|}{0.255} & \multicolumn{1}{l|}{0.444} & \multicolumn{1}{l|}{0.860} & 0.925 & -0.110    \\ 
                             \cline{2-8}
                             
                             & $\sigma_{s.d.}$ & \multicolumn{1}{l|}{0.028} & \multicolumn{1}{l|}{0.136} & \multicolumn{1}{l|}{0.189} & \multicolumn{1}{l|}{0.050} & 0.028 & 0.013 \\ 
                             \hline
    \end{tabular}
    \end{table}

        \begin{figure}[!ht]
        \centering
        \includegraphics[scale=0.75]{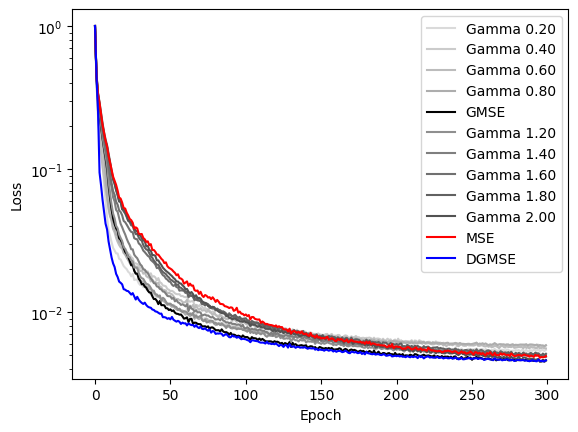}
        \caption{Generator validation loss for the baseline MSE loss optimised network (red) and the GMSE loss gamma variant networks. More rapid convergence is observed for the GMSE optimised network, in comparison to the MSE optimised network. The DGMSE optimised network is observed to outperform both approaches.}
        \label{fig:Gen_Loss2}
        \end{figure}

        \begin{figure}[!ht]
        \centering
        \includegraphics[scale=0.750]{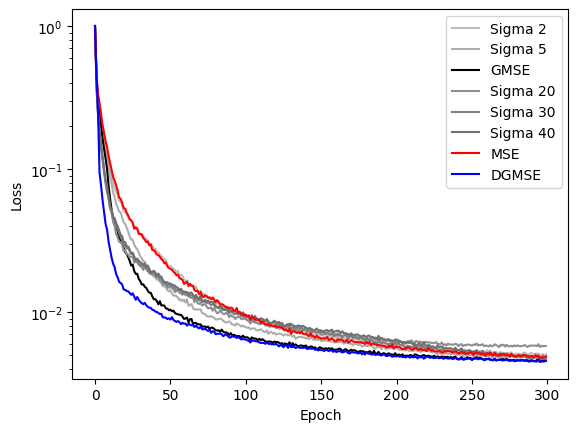}
        \caption{Generator validation loss for the baseline MSE loss optimised network (red) and the GMSE loss sigma variant networks.}
        \label{fig:Gen_Loss3}
        \end{figure}

        \begin{figure}[!ht]
        \centering
        \includegraphics[scale=0.750]{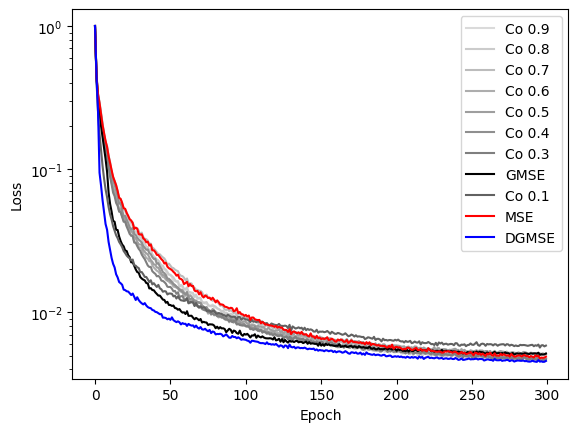}
        \caption{Generator validation loss for the baseline MSE loss optimised network (red) and the GMSE loss offset variant networks.}
        \label{fig:Gen_Loss4}
        \end{figure}

        \begin{figure}[!ht]
        \centering
        \includegraphics[scale=0.750]{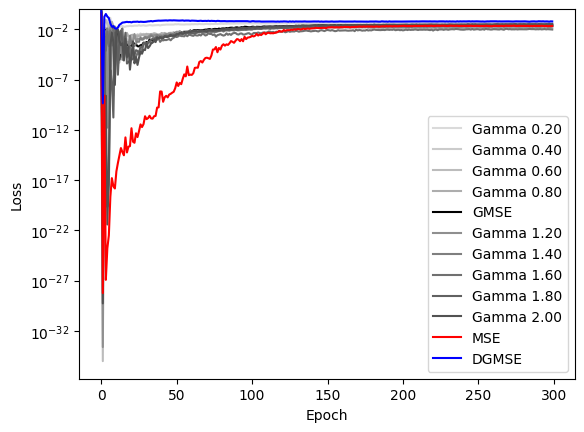}
        \caption{Discriminator loss on generated images for the baseline MSE loss optimised network (red) and the GMSE loss gamma variant networks.}
        \label{fig:Disc_Loss2}
        \end{figure}

        \begin{figure}[!ht]
        \centering
        \includegraphics[scale=0.750]{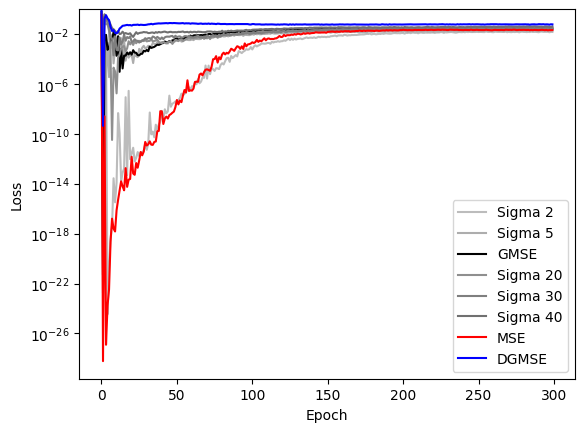}
        \caption{Discriminator loss on generated images for the baseline MSE loss optimised network (red) and the GMSE loss sigma variant networks.}
        \label{fig:Disc_Loss3}
        \end{figure}

        \begin{figure}[!ht]
        \centering
        \includegraphics[scale=0.750]{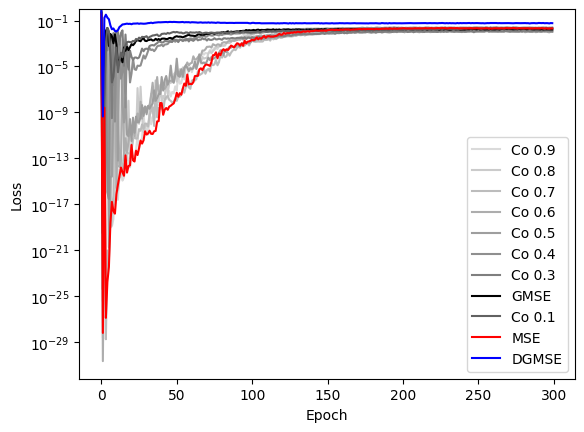}
        \caption{Discriminator loss on generated images for the baseline MSE loss optimised network (red) and the GMSE loss offset variant networks.}
        \label{fig:Disc_Loss4}
        \end{figure}

        Three plots detail the generator loss during training. The Adam optimiser has been used to reduce this loss during training for each individual test. The maximum loss values have been normalised to 1 for all data shown. Fig. \ref{fig:Gen_Loss2} provides the gamma ($\gamma$) generator loss, Fig. \ref{fig:Gen_Loss3} provides the sigma ($\sigma$) generator loss and Fig. \ref{fig:Gen_Loss4} provides the offset loss ($C_{o}$) generator loss. The x axis details the epochs of training, and the y axis provides the normalised loss of each test. Indicated in red is the MSE baseline, black details the GMSE baseline and blue the DGMSE loss. Three additional figures are provided, covering the discriminator loss during training. A separate CNN network was used to assess both the real and generated images during training. The goal of the generator was to produce images that would be awarded a value of 1, indicating the generated fake image belonged to the real dataset when scored by the discriminator. Fig. \ref{fig:Disc_Loss2} provides the gamma variant discriminator performance, Fig. \ref{fig:Disc_Loss3} provides the sigma variant and Fig. \ref{fig:Disc_Loss4} the offset variant performance. The y axis shows the score from the discriminator where 1 is the target and the x axis denotes the epoch of training. Again, indicated in red is the MSE baseline, black is the GMSE baseline and blue the DGMSE loss.  

\begin{figure}[H]
        \centering
        \includegraphics[width=\textwidth]{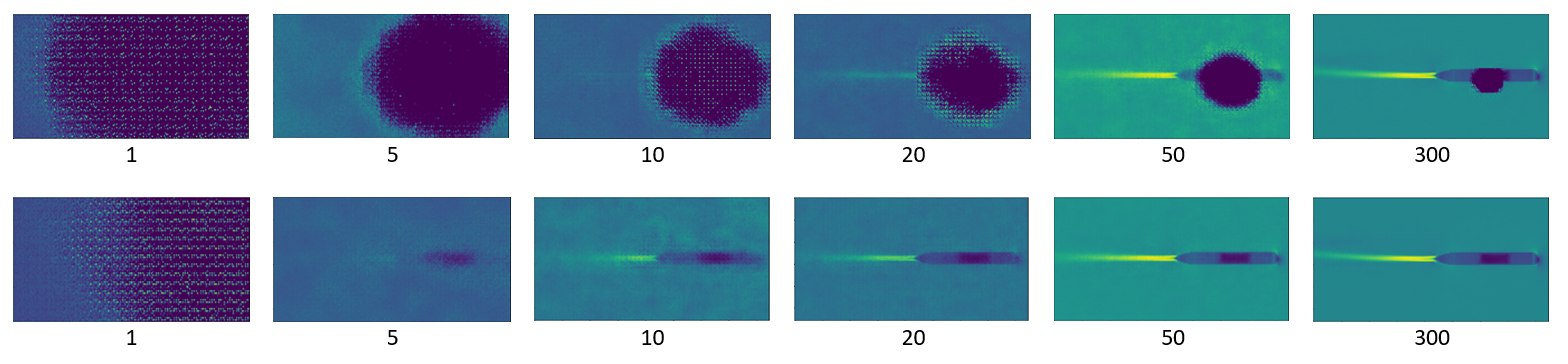}
        \caption{Generated image quality during training presented across training. Top row is the MSE baseline optimised network, bottom row is the GMSE baseline ($\sigma = 10, \gamma = 1.00$ and $C_{o} = 0.2$) optimised network. A different dataset instance to Fig. \ref{fig:GroundTruth} is used to provide the reconstruction where stronger gradients are present.}
        \label{fig:Training_images}
        \end{figure}

        The generator loss of the gamma variations provided in Fig. \ref{fig:Gen_Loss2} were seen to converge faster than the MSE loss optimised network. Both the GMSE and DGMSE loss functions provided beneficial loss characteristics in comparison to the gamma variants themselves, which were seen to converge at different rates, but to a slower extent than the GMSE baseline (black). Initial rates of loss in the gamma 0.20, 0.40 and 0.60 were seen to outperform the GMSE baseline to a limited extent, but with ultimately lower maximum rates, as reflected by Table \ref{tab:data_table}. Gamma variations of 0.20 to 0.80 were seen to converge to a higher final loss than $\gamma >= 1.00$ (Fig. \ref{fig:Gen_Loss2}), indicating that the loss function may benefit from dynamic changes to $\gamma$ during training. The dynamic variant, DGMSE, was seen to outperform all gamma variants and both the MSE and GMSE baselines and resulted in the lowest final loss. The ability of the gamma variants to fool a discriminator network, Fig. \ref{fig:Disc_Loss2}, is seen to have limited variation and relatively robust performance compared to the sigma discriminator loss (Fig. \ref{fig:Disc_Loss3}) and the offset discriminator loss (Fig. \ref{fig:Disc_Loss4}). All gamma variants are observed to fool the discriminator network, at earlier epochs, compared to the MSE baseline. $\gamma > 1.00$ variations are seen to have better performance at early epochs ($< 50$), but worse performance later in training ($ > 50$). The GMSE baseline, with no $\gamma$ shift was observed to perform robustly, with moderate performance in early training but a better ability to overcome the discriminator at later stages. The DGMSE loss was observed to perform well early in training and until completion, where it recorded the highest final loss. The second highest final loss was the GMSE loss function, followed by the $0.20 < \gamma =< 1.00$ variants, then the MSE loss and finally the $1.20 < \gamma =< 2.00$ variants. Thus, further supporting the adoption of a dynamic approach (DGMSE) or $\gamma =< 1.00$. 
\\ \\
        Sigma was seen to also effect the generator loss during training (Fig. \ref{fig:Gen_Loss3}), but to differing extents depending on the strength of the blur. In the initial stages of training ($< 50$ epochs), strong blurs ($\sigma > 10$) corresponded to better loss rates. As training progressed however ($> 50$ epochs) smaller blurs became more performant and resulted in lower final loss values at the completion of training. The MSE baseline was found to outperform the sigma 2 and 5 variants, only at the completion of training. Both the GMSE and DGMSE outperformed the baselines and all sigma variants, with the DGMSE displaying beneficial initial loss, where it outperformed all variants with the highest recorded normalised loss rate (-0.189). The ability of the sigma variants to fool the discriminator (Fig. \ref{fig:Disc_Loss3}) is also dependent on blur strength. Low strength blurs ($\sigma = 2$) are observed to perform similarly to the MSE baseline, whilst stronger blurs offer a discriminator loss advantage with markedly better early perform loss (epochs $< 100$), after which the strength of the blur is less impactful and closer loss performance is observed. Again, the DGMSE is the most performant, followed by the sigma variants in descending order of strength (40 to 20), then the GMSE variant, sigma 5 variant, the MSE baseline and finally the sigma 2 variant. 
\\ \\        
        The offset weighting of the gradient free region was also seen to have an effect on generator loss convergence, as provided in Fig. \ref{fig:Gen_Loss4}. Lower values indicate where the gradient region is highly important and higher values prioritise the gradient free region. Initial convergence of all offset variants outperformed the MSE baseline, with this continuing for the entirety of training, with $C_{o} = 0.1$ a notable exception to this, as it resulted in higher loss from 130 epochs onwards. The $C_{o} = 0.2$ variant was also found to result in higher loss than the MSE baseline at the completion of training. These findings indicate that both the gradient and gradient free regions retain importance during training and the weighting should account for both. The higher performance of the GMSE and DGMSE baselines indicate the benefits of either a balanced or dynamic approach. For the offset discriminator loss performance (Fig. \ref{fig:Disc_Loss4}), offsets equal to or above 0.5 are seen to perform close to the MSE baseline in early training between 0 and 100 epochs, before converging to lower loss values (indicating decreased performance). Conversely, offsets less than 0.5, where the gradient regions are weighted heavily, correspond to better initial training (0 to 100 epochs) as well as better final convergence. The most beneficial final loss is achieved by the DGMSE variant, followed by $C_{o} = 0.1$, then the GMSE baseline (with $C_{o} = 0.1$), then $0.3 < C_{o} < 0.5$, the MSE baseline and then $0.6 < C_{o} < 0.9$.   
\\ \\
        Qualitative field assessment also offers insight into the behaviour of the baseline MSE and GMSE loss optimised networks. Fig. \ref{fig:Training_images} shows the generation abilities of the two baseline networks as training progressed. Both networks have similar performance after a single epoch, before deviation at the displayed 5, 10, 20, 50 and 300 epochs. It was seen that a circular spatial reconstruction error was present where the payload bay of the submarine was located in the fluid flows of \cite{ZacIEEE}. This error does not appear in the GMSE loss optimised network, where these region had a higher weighting for accurate generation. The strong gradient regions around the submarine developed earlier in the training process for the GMSE loss optimised network and were refined at later stages of training.
            
\section{Discussion} \label{Discussion2}

The results of these tests demonstrate the potential of gradient capture driven GMSE and DGMSE loss functions for accelerating training speeds of generative architectures and ensuring higher quality generations for fluidic applications. The dynamic extraction and weighting of regions of importance was demonstrated to be a robust means of accelerating loss convergence at both the generator and discriminator network of a cGAN architecture, resulting in several advantages when compared to MSE loss. These advantages are evident in several aspects. The first is the structural similarity index measure (SSIM) performance which was observed to consistently surpass that of the MSE baseline across all variants of the GMSE loss function. The higher recorded SSIM indicates a more accurate reconstruction of the complex gradient structures of the flow field, as evidenced by both Fig. \ref{fig:Training_images} and the data of Table \ref{tab:data_table}. The 82.1\% and 83.6\% reduction in relative SSIM error achieved by the GMSE and DGMSE loss function highly the potential of leveraging gradient information to improve generation fidelity.
\\ \\
Notably, the DGMSE and GMSE variants resulted in a faster rate of convergence during training, evidenced by the higher maximum loss rates in comparison to the MSE baseline. This accelerated learning can be attributed to the GMSE function’s ability to provide more informative gradients to the generator, allowing for faster and more effective optimisation. This finding carries significant practical implications, as faster training directly translates to reduced computational cost and faster development cycles for deep learning. Furthermore, the DGMSE and GMSE loss functions success in consistently and rapidly fooling the discriminator network highlights its capacity to guide the generator towards producing more realistic and challenging flow fields for the discriminator to score. This superior performance, particularly at early epochs (Fig. \ref{fig:Disc_Loss2}, Fig. \ref{fig:Disc_Loss3}, Fig. \ref{fig:Disc_Loss4}) suggests that the GMSE loss facilities a more efficient exploration of the data and results in a more effective approximation of the underlying data distribution of the RANS dataset.
\\ \\
Exploration of the GMSE function's parameters reveals their influence on its efficacy. The blur strength, controlled by $\sigma$, highlights a crucial trade-off. Weaker blurs ($\sigma < 10$) retain precise gradient localization but risk overfitting to small-scale features, while stronger blurs ($\sigma > 20$) provide a more generalized weighting, potentially diluting the focus on crucial regions, but are still seen to outperform MSE loss. The optimal blur strength is likely dataset-dependent, but DGMSE substantiates the benefits of starting with a strong blur and decreasing the strength during training. This technique is seen to result in beneficial early and late stage SSIM, as well as a higher maximum loss rate (Fig. \ref{fig:Gen_Loss3}). 
\\ \\
The gamma parameter, $\gamma$, governs the contrast enhancement of the weighting mask. While its overall impact on SSIM appears limited within the tested range, it significantly affects the convergence behaviour (Table \ref{tab:data_table}). Higher gamma values ($\gamma > 1$), emphasizing high-gradient regions, lead to faster initial convergence but potentially at the cost of a higher final loss. This suggests a dynamic gamma adjustment strategy could further optimize training by initially prioritizing rapid error reduction and later shifting focus to fine-grained details.
\\ \\
The offset parameter, $C_o$, dictates the relative weighting of gradient-rich versus gradient-free regions. Assigning high weights to gradient-free regions ($C_o >= 0.4$) consistently degraded performance, highlighting the importance of prioritizing areas containing crucial flow structures. Conversely, excessively low offset values ($C_o = 0.1$) might overshadow the contribution of the surrounding context. The optimal offset likely depends on the specific dataset and the relative significance of high-gradient features versus their surroundings.
\\ \\
The DGMSE and GMSE loss functions are also seen to achieve these advantages without requiring domain specific knowledge or complex modifications to the training data, to account for highly customised loss functions \cite{CFD_MSE_apriori}. While these customised loss functions for CFD data do exist, they introduce significant complexity \cite{CFD_LossFunction} and can lead to unstable or incomparable training \cite{CFD_LossFunction} – limiting their generalizability. The DGMSE and GMSE loss function, in contrast, relies on the simpler method of prioritizing the regions of high gradient disparity – making it easily adaptable to other datasets exhibiting similar characteristics – such as temperature fields \cite{Temp_Data}, hydrodynamic \cite{Hydro_Data} and aerodynamic data \cite{Aero_Data} and fields with strong edge structures \cite{MNIST_Data}.
\\ \\
Generational architectures are susceptible to a range of different failure modes or issues that render training challenging. Most often discussed are mode collapse and convergence failure \cite{ModeCollapse1}. Mode collapse occurs where the generator struggles to produce a variety of data and produces one output for a larger range of inputs \cite{ModeCollapse2}. This is prevalent when working with small datasets and model sizes, where either the model or dataset is not sufficiently large, and struggles to approximate the underlying data distribution. This often leads to generative errors, in the case of GAN architectures. In a limited capacity, this can be seen in Fig. \ref{fig:Training_images} where the centre of the submarine in the MSE images (top) can be seen to feature a consistently low field generation, despite this not being the desired output. This can be contrast against the more correct GMSE generation (bottom). The effective prevention of mode collapse is challenging but may be alleviated by the function of the GMSE loss function. Giving an increased weighting to the images regions of importance in the GMSE loss function results in the generator being proportionally punished for generating poor outputs, in areas such as the submarine itself, leading to a more robust generative ability. Using the natural weighting behaviour of the loss function to combat mode collapse offers an easier pathway to prevention than the implementation of other, more complex methods of prevention, like those explored in literature \cite{ModeCollapse3}. This is because the generator is incentivised to produce higher quality images by way of the weighted loss term.

\section{Conclusions} \label{Conclusion}

This research introduces the GMSE loss function as a powerful and versatile tool for enhancing the training of generative deep learning architectures, with a focus on fluid fields and gradient-drive data. The ability of both the DGMSE and GMSE loss function to dynamically prioritise gradient containing regions, whilst remaining conceptually simple and adaptable underscores its potential for broader applicability. Future research should investigate the performance of the GMSE loss function with other gradient-containing datasets, explore its application to higher dimension data (such as 3D volumes) and delve into its potential for mitigating mode collapse in generative models. This work represents a significant step towards efficient, effective and generally applicable training techniques for generative architectures across a range of fields. 

\section*{CRediT authorship contribution statement}
\textbf{Zachary Cooper-Baldock}: Investigation, Conceptualization, Methodology, Formal analysis, Writing - original draft, Writing - review and editing. \textbf{Paulo E. Santos, Russell S.A. Brinkworth, Karl Sammut}: Supervision, Conceptualization, Methodology, Writing - review and editing. 

\section*{Declaration of competing interest}
The authors declare that they have no known competing financial interests or personal relationships that could have appeared to influence the work reported in this paper.

\section*{Data availability}
Data will be made available upon reasonable request.

\section*{Acknowledgments}
\noindent This work was supported by an NCI HPC-AI Talent Program 2023 Scholarship, with computational resources provided by NCI Australia, an NCRIS enabled capability supported by the Australian Government.

\appendix
\section{Architectures}
\label{app1}

\begin{figure}[H]
        \centering
        \includegraphics[scale=0.60]{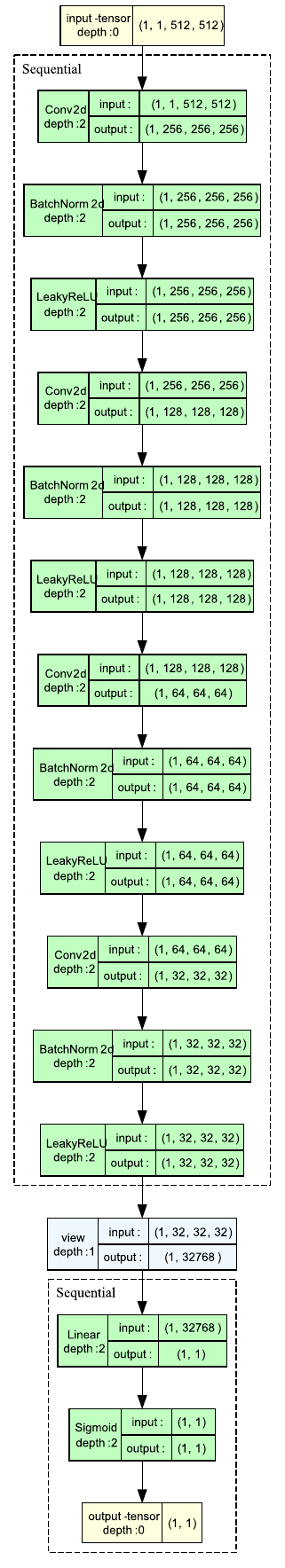}
        \caption{Pytorch architecture for the discriminator network. Layer structure, activation functions and inputs as shown. Sequential regions denoted by green body structure. Input and output denoted by yellow cells.}
        \label{fig:image1}
        \end{figure}

\begin{figure}[H]
        \centering
        \includegraphics[scale=0.75]{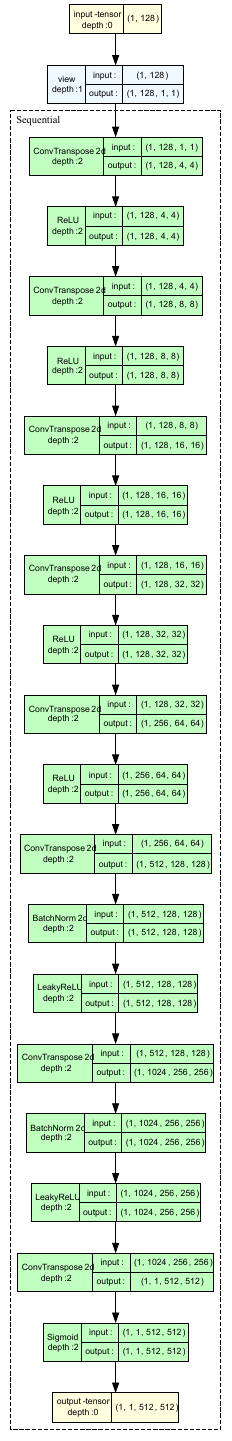}
        \caption{Pytorch architecture for the Generator network. Layer structure, activation functions and inputs as shown. Sequential regions denoted by green body structure. Input and output denoted by yellow cells.}
        \label{fig:image2}
        \end{figure}

Contained here are the architecture schematics of the controlled generative adversarial network used in the investigation of the GMSE and DGMSE loss functions. Fig. \ref{fig:image1} displays the discriminator architecture and Fig. \ref{fig:image2} displays the generator architecture. 

\bibliographystyle{elsarticle-num} 
\bibliography{Reference_List}

\end{document}